\documentclass[letterpaper, 10 pt, conference]{ieeeconf}  % Comment this line out if you need a4paper

\IEEEoverridecommandlockouts                              % This command is only needed if 
\let\labelindent\relax

\usepackage{amsmath} % assumes amsmath package installed
\usepackage{amssymb}  % assumes amsmath package installed
\usepackage{gensymb}
\usepackage{comment}
\usepackage{graphicx}
\usepackage{caption}
\usepackage{subcaption}
\usepackage{xspace}
\usepackage{tabularray}
\usepackage{arydshln}
\usepackage{multirow}
\usepackage{caption}
\usepackage{flushend}
\usepackage[inline]{enumitem}
\makeatletter
\let\NAT@parse\undefined
\makeatother
\usepackage{hyperref}

\hypersetup{
    colorlinks=true,
    }

\usepackage{xcolor}
\definecolor{patrick_color}{rgb}{.0,.6,.05}
\definecolor{mengyu_color}{rgb}{.5,.7,.1}
\definecolor{samarth_color}{rgb}{0,0.35,0}
\definecolor{arun_color}{rgb}{0,0,1}
\definecolor{charlie_color}{rgb}{0,0,0.8}
\definecolor{james_color}{rgb}{0.75,0.25,0.0}

\makeatletter
\DeclareRobustCommand\onedot{\futurelet\@let@token\@onedot}
\def\@onedot{\ifx\@let@token.\else.\null\fi\xspace}

\def\eg{\emph{e.g}\onedot}

\makeatother

\newcommand{\datasetName}{{Robot Kidnapper}\xspace}

\title{\LARGE \bf
The Un-Kidnappable Robot: Acoustic Localization of Sneaking People
}

\author{Mengyu Yang$^{1*}$, Patrick Grady$^{1}$, Samarth Brahmbhatt$^{2}$,\\ Arun Balajee Vasudevan$^{3}$, Charles C. Kemp$^{4}$, James Hays$^{1}$% <-this % stops a space
\thanks{$^{*}$Corresponding author: 
        {\tt\small my.yang@gatech.edu}}%
\thanks{$^{1}$Mengyu Yang, Patrick Grady, and James Hays are with the Georgia Institute of Technology (Georgia Tech)}%
\thanks{$^{2}$Samarth Brahmbhatt is with Intel Labs}%
\thanks{$^{3}$Arun Balajee Vasudevan is with Carnegie Mellon University}%
\thanks{$^{4}$Charles C. Kemp worked for Georgia Tech and now works full-time for Hello Robot Inc., which sells the Stretch RE-1}%
}

\begin{document}

\maketitle
\thispagestyle{empty}
\pagestyle{empty}

%%%%%%%%%%%%%%%%%%%%%%%%%%%%%%%%%%%%%%%%%%%%%%%%%%%%%%%%%%%%%%%%%%%%%%%%%%%%%%%%

\begin{abstract}

How easy is it to sneak up on a robot? We examine whether we can detect people using only the incidental sounds they produce as they move, even when they try to be quiet. To do so, we first collect a robotic dataset of high-quality 4-channel audio paired with 360\degree{} RGB data of people moving in different indoor settings. Using this dataset, we train models to predict if there is a moving person nearby and then their location using only audio. We implement our method on a robot, allowing it to track a single person moving quietly using only passive audio sensing. For demonstration videos, see our~\href{https://sites.google.com/view/unkidnappable-robot}{project page}.
\end{abstract}

%%%%%%%%%%%%%%%%%%%%%%%%%%%%%%%%%%%%%%%%%%%%%%%%%%%%%%%%%%%%%%%%%%%%%%%%%%%%%%%%
% \begin{figure}[t!]
%     \centering
%     \begin{subfigure}[b]{0.48\columnwidth}
%         \centering
%         \includegraphics[width=\columnwidth]{figures/dataset third person.png}
%         \caption{The \datasetName dataset}
%         \label{fig:dataset third person}
%     \end{subfigure}
%     \begin{subfigure}[b]{0.48\columnwidth}
%         \centering
%         \includegraphics[width=\columnwidth]{figures/robot_demo.png}
%         \caption{Person detection}
%         \label{fig:demo}
%     \end{subfigure}
%     \caption{Can we detect where people are based only on the subtle sounds they incidentally produce when they move? (a) We collect a dataset of high-quality audio paired with 360\degree{} RGB data with multiple participants in different indoor scenes. (b) We train models to perform audio-based person detection which can be implemented on a robot in real time.}
%     \label{fig: hardware setup}
% \end{figure}

\section{INTRODUCTION} \label{sec:intro}

Advances in mobile robots have led to such platforms becoming increasingly common in everyday settings. With this popularity comes a rise in the coexistence of robots with people. Nowadays, it is not uncommon to see a last-mile delivery robot roaming a city sidewalk, an industrial robot navigating a warehouse floor, or a cleaning robot vacuuming in a home. As demand for robotic applications grows, being able to recognize people in the robots' proximity is a vital task to ensure safety. Object recognition in general has been well examined for image data~\cite{ren2015faster,he2017mask, carion2020end}, where humans are  one of the object categories. Previous works have also specifically investigated person detection~\cite{Linder2021CrossModalAO, jia2021domain, jia2020dr, yan2018multisensor,fung2023robots} by detecting the presence of people as well as localizing them. The large amount of existing research in this topic highlights the universality of person detection across different robotic applications. 

\begin{figure}
    \centering
    \includegraphics[width=\columnwidth]{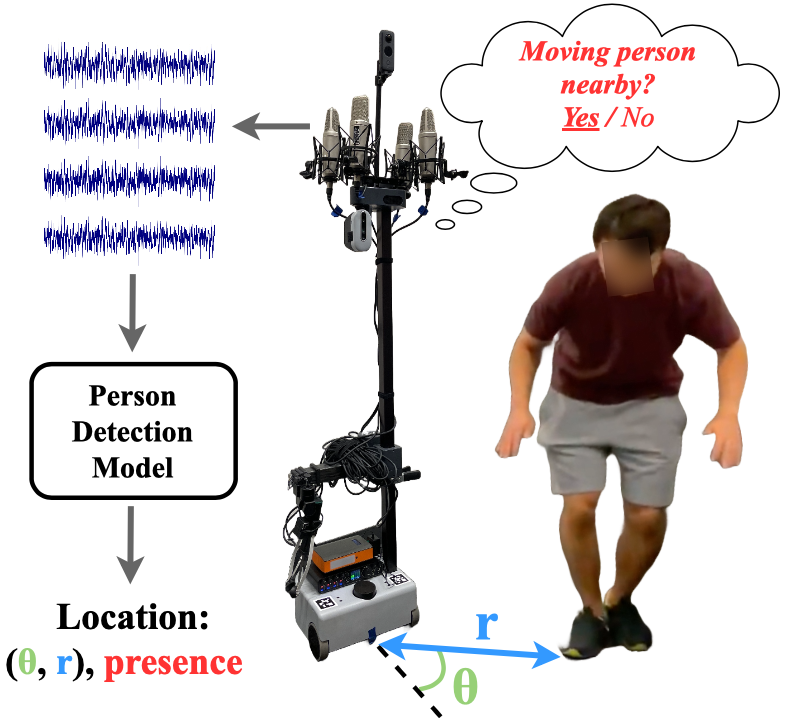}
    \caption{Can we detect where people are based only on the subtle sounds they incidentally produce when they move, even when they try to be quiet? We collect a dataset of high-quality audio paired with 360\degree{} RGB data with different participants in multiple indoor scenes. We train models to localize a moving person based on audio only and implement it on a robot.}
    \label{fig:teaser}
\end{figure}

\begin{figure*}
    \centering
    \includegraphics[width=0.98\textwidth]{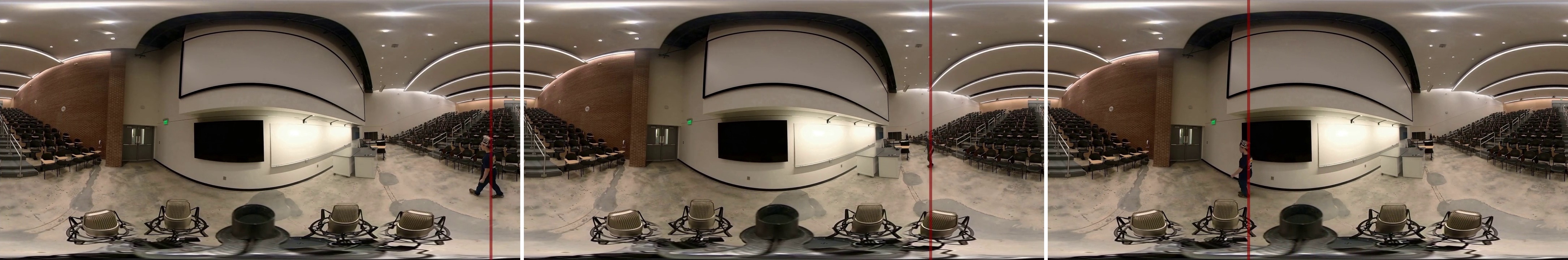}
    \caption{Frames from the \datasetName dataset (static robot). The participant wears a hat with ArUco markers~\cite{garrido2014automatic} used to calculate ground truth radial distance. The RGB frames are used to calculate the ground truth centroid of the person using DeepLabv3+~\cite{chen2018encoder}. Only the audio is used during training. The vertical red lines are the angles predicted by our model in an unseen room. The participant is walking normally in these frames.}
    \label{fig:dataset frames}
\end{figure*}

Most person detection methods use vision- and spatial-based sensors. These include RGB~\cite{Linder2018TowardsA3}, depth~\cite{Linder2020accurate}, 2D lasers~\cite{beyer2018deep,2021arXiv210611239J}, 3D LIDAR~\cite{yan2018multisensor}, or combinations for multi-modal methods~\cite{ku2018joint, linder2016on}. While multi-modal models are beneficial for domain adaptation and improved performance over uni-modals, these methods learn a joint representation across all sensors that do not allow flexibility for variable sensor inputs. These models are therefore not robust against failures in real-world applications where sensors can fail. Aside from these common sensors, fewer works have examined the use of audio for person detection. Existing works only use audio to estimate the direction of arrival (DOA)~\cite{singh2023sporadic,yamazaki2020audio, Ciuffreda2023people, sewtz2020robust} which does not satisfy our definition of person detection. Others that do perform full localization tend to rely on easily detectable sounds like talking or audio from a loudspeaker~\cite{portello2011acoustic, bechler2004system, shi2023audio, Gala2018RealtimeAS}.

We argue that the subtle acoustics \emph{incidentally} produced by people as they move around are an under-leveraged source of information that can be used for person detection. The incidental sounds we focus on are by nature noisy and weak. We demonstrate this by showing how other sound localization methods~\cite{knapp1976generalized, chen2022sound} often fail to detect the sound source for this type of audio. Unlike other sound localization methods, ours relies on \textit{passive observation} only. We only record the sounds already coming from the environment and do not produce any additional sounds to aid in detection. This is in contrast to other methods using active sensing, such as with ultrasonic sensors~\cite{Tripathi2021human}, echolocation~\cite{Christensen2019BatVisionLT}, or the impulse response of a room~\cite{wang2023soundcam}.

Having an audio-only based method for person detection is an important step in the development of multi-modal person detection systems that are robust to failures. Should the sensors that many frameworks rely on fail or become unavailable (low-lit environments, occlusion handling, etc.), our method allows robots to fall back solely onto audio, a readily obtainable signal which is usually already onboard most hardware setups. And when interacting with robots, people should not be expected to intentionally create extra sounds to ensure nearby robots are aware of their location.  

To evaluate our claims, we first collected a real-world dataset of different people moving around a robot in various indoor settings. Onboard the robot, we record 4-channel, high-quality audio along with paired 360\degree{} RGB data, which we process to obtain pseudo-labels for the person's location relative to the robot. We name this the \datasetName dataset (Fig.~\ref{fig:dataset frames},~\ref{fig:dataset third person}) and provide more details in Section~\ref{sec: Dataset}.

We then use this dataset to learn person detection based on the incidental and often subtle sounds created by people as they move around. We show that our models are able to localize people both when the robot is stationary and when the robot is moving, a more difficult task due to the additional self-noise of the robot. We also implement our model on a real robot to demonstrate robotic human awareness using only audio. Overall, we present the following contributions: 

\begin{enumerate*}[label=\textbf{\arabic*})]
    \item A public dataset of synchronized, high-quality, 4-channel audio and 360\degree{} RGB data of different participants in multiple indoor scenes.

    \item Experimental evaluations of person detection using the subtle, incidental sounds of people moving around.

    \item Allowing real robots to track people using only the sounds of them moving.
\end{enumerate*}

\section{RELATED WORK}
\subsection{Human Detection with Visual Perception}
Since perceiving humans enables a large number of downstream tasks, `person' or `human' is included as a category in most image-based object detection~\cite{tan2020efficientdet, zhou2022detecting} and segmentation~\cite{kirillov2023segment} models. Autonomous cars use LIDAR sensors optionally combined with RGB cameras to detect pedestrians~\cite{premebida2014pedestrian, navarro2016machine} and forecast their future behaviour~\cite{wilson2021argoverse}. Additionally, surveillance systems use RGB or infrared images to detect~\cite{yu2020scale, park2019cnn, zin2011fusion}, identify~\cite{ye2021deep}, and localize humans~\cite{ristani2018features} and objects~\cite{object2015occlusion}. In settings where it can be assumed that any disturbances are caused by humans, human detection is performed through anomaly detection algorithms. For example, laser/ultraviolet beam breakers or proximity sensors are used on factory floor automated assembly lines~\cite{ceriani2013optimal}, while some intrusion detection security systems~\cite{zieger2009acoustic} use audio. In contrast to these works, our paper focuses solely on passively sensing audio signals to not only detect but also localize humans.

% \subsection{Robot Person Following}

\subsection{Audio-Based Perception for Robots}
Audio has been used for robotic tasks involving both non-human and human interactions such as self-localization~\cite{chu2006scene, hornstein2006sound}, robotic pouring~\cite{liang2019making}, and navigation using ambient sounds~\cite{chen2021structure}. Sound source localization systems like~\cite{valin2003robust, wilson2020avot, sasaki2018online} usually assume that the source emits loud, obvious sounds (\eg beeps, music, speech). But as Fig.~\ref{fig:spectrograms} shows, the more subtle sounds we focus on have different acoustic characteristics which these algorithms may not successfully generalize to. \cite{singh2023sporadic, yamazaki2020audio, Ciuffreda2023people, sewtz2020robust} all examine human detection but only estimates the direction of arrival. In contrast, we perform full 2D localization by also predicting the radial distance of the person from the robot.
% A common human-based task involves human detection, but these works tend to only use audio to estimate the direction of arrival of humans~\cite{singh2023sporadic, yamazaki2020audio, Ciuffreda2023people, sewtz2020robust}. In contrast, we focus on detection \textit{and} localization of a person using much quieter sounds, like those incidentally made by a person trying to walk quietly. 
And while Sasaki et al.~\cite{sasaki2018online} performs 3D localization, they assume a static sound source with a moving robot. In our work, we assume both the robot and sound source can be moving at the same time.

% Audio information has also been used for robot tasks not involving human interaction \eg , object or sound source localization~\cite{valin2003robust, liang2019making, wilson2020avot}, and object state perception~\cite{liang2019making}.
% Sasaki et al.~\cite{sasaki2018online} localizes the robot using a microphone array hearing to multiple static speakers. We differ from their work in localizing incidental sounds of people who can be in motion.

\section{Dataset} \label{sec: Dataset}

To train our models to detect people based on the incidental sounds that they produce, we collected the \datasetName{} dataset. This dataset contains high-quality 4-channel audio recordings paired with 360\degree{} RGB video from the robot's egocentric point of view (Fig.~\ref{fig:dataset frames}). The person's position was annotated in coordinates relative to the robot. 
%(detailed in Section\ref{sec: Dataset} B).% 
We collected data in 8 rooms across 4 buildings. To account for the potential impacts that physical properties of a room may have, the selected rooms vary in terms of size (small study room, large lecture hall, etc.) and material (concrete floors, carpeted floors, glass walls, etc.).

\subsection{Human Presence Recordings} 

The \datasetName{} dataset captures 12 participants in a range of environments performing a variety of actions. This \emph{stress-tests} the performance of our algorithm across diverse behaviors. Participants were prompted to perform 4 different actions during data capture to capture a wide range of sounds:

\begin{itemize}
    \item \emph{Stand still:} Participants were asked to stand in place for 5 seconds before taking 1-2 steps to a different spot and repeating the procedure.
    \item \emph{Walk quietly:} Participants were prompted to move during the entire recording but to focus on minimizing any sounds that they produced.  %procedure. \patrick{I renamed this from sneaky since IMO sneaky doesn't sound academic enough}
    \item \emph{Walk normally:} Participants were prompted to walk at their normal speed and volume
    \item \emph{Walk loudly:} Here, participants were prompted to walk more loudly, which they accomplished by dragging their feet or stomping.
\end{itemize}

During data collection, the hardware was mounted to a Stretch RE-1 mobile manipulator robot. However, the robot produces sounds during movement, such as humming and clicking from wheel motors or rustling as the robot traverses bumps on the ground. We examine whether our methods can learn under the more difficult setting of detecting a moving person with these additional noises. For all 4 actions, we collected data under a \textit{static robot} and \textit{dynamic robot} condition. During the \textit{static} recordings, the robot was turned on but remained stationary. In the \textit{dynamic} recordings, the robot was teleoperated from outside the room and driven (translation and rotation) along a random path, with the aim of uniformly covering the entirety of the room. The robot moved at a translational and rotational velocity of approximately 0.25 m/s and 0.17 rad/s, respectively. Participants were alone in the room with the robot. As our intention is to examine how well our models can detect people from only incidental sounds, non-incidental sounds such as talking were cropped out during post-processing. All recordings contain a single participant only.  The data collection was approved by an Institutional Review Board (IRB) and participants gave informed consent and were compensated for their time. All combined, the human presence recordings total to approximately 8 hours, evenly split between all actions, robot conditions, and rooms. 

\subsection{Empty Room Recordings} \label{sec: empty room}

In addition to recording the sounds of human presence, we also recorded audio of the 8 rooms used in the dataset when it was empty. This empty room data helps our model be able to distinguish whether or not there is a moving person in the robot's vicinity. The empty room recordings are collected on the same robot setup as the human recordings. It is also split between \textit{static} and \textit{dynamic}. This empty room dataset is approximately 5 hours in length.   
% \subsection{Empty Room Augmentation Dataset} \label{sec:aug dataset}
We then collected a secondary empty room dataset without the Stretch RE-1 consisting only of short recordings. The \textit{Empty Augmentation} dataset was collected in 26 rooms across 6 buildings on Georgia Tech's campus. In each room, audio of the empty room was recorded from 2 different positions for 2 minutes each, resulting in around 1.5 total hours of audio. The audio from this dataset is used for data augmentation, which is described in Section~\ref{sec:empty aug}.

\subsection{Person Location Labels}

Training a person detection model requires ground truth labels for the location of the person relative to the robot. In our dataset, we annotate the person's position on the ground plane in polar coordinates. Specifically, we annotate the azimuthal angle $\theta$ of the person relative to the robot's forward vector, and radial distance $r$ of the person relative to the robot's origin. To label $\theta$, we developed an approach based on a semantic segmentation model. Specifically, we use a pre-trained DeepLabv3+~\cite{chen2018encoder} model to generate a mask of the person from an RGB frame and calculate the centroid of the person $(x, y)$. We then encode $x$ using cyclical features, where $W$ is the width of the frame. 
\begin{gather}
    \theta_{sin} = sin(2\pi x / W)\\
    \theta_{cos} = cos(2\pi x / W)
\end{gather}

% Rather than hand-labelling our dataset, we use a pre-trained DeepLabv3+~\cite{chen2018encoder} semantic segmentation model. We detect the segmentation map of only the person class and calculate its centroid $(x, y)$. We use equirectangular projection for the 360\degree{} frames, meaning the width of the image maps to the entire 360\degree{} of the captured frame (Fig. *). We then encode $x$ using cyclical features with $\{\theta_{sin} = sin(2\pi x / W), \theta_{cos} = cos(2\pi x / W)\}$, where $W$ is the width of the image. 

\begin{figure}[t]
    \centering
    \begin{subfigure}[b]{0.36\columnwidth}
        \centering
        \includegraphics[width=\columnwidth]{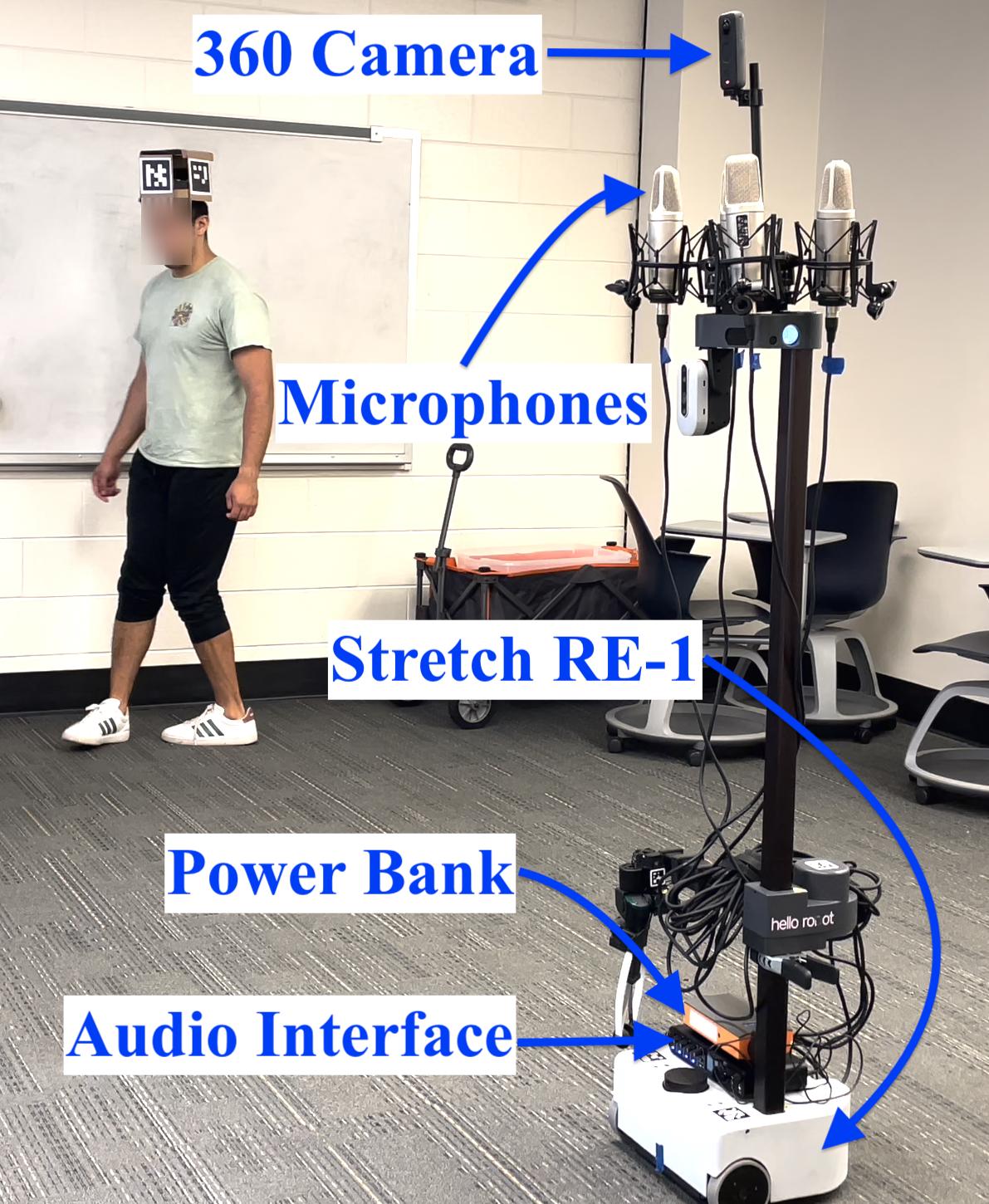}
        \caption{}
        \label{fig:dataset third person}
    \end{subfigure}
    % \hspace{0.0\columnwidth}
    \begin{subfigure}[b]{0.56\columnwidth}
        \centering
        \includegraphics[width=\columnwidth]{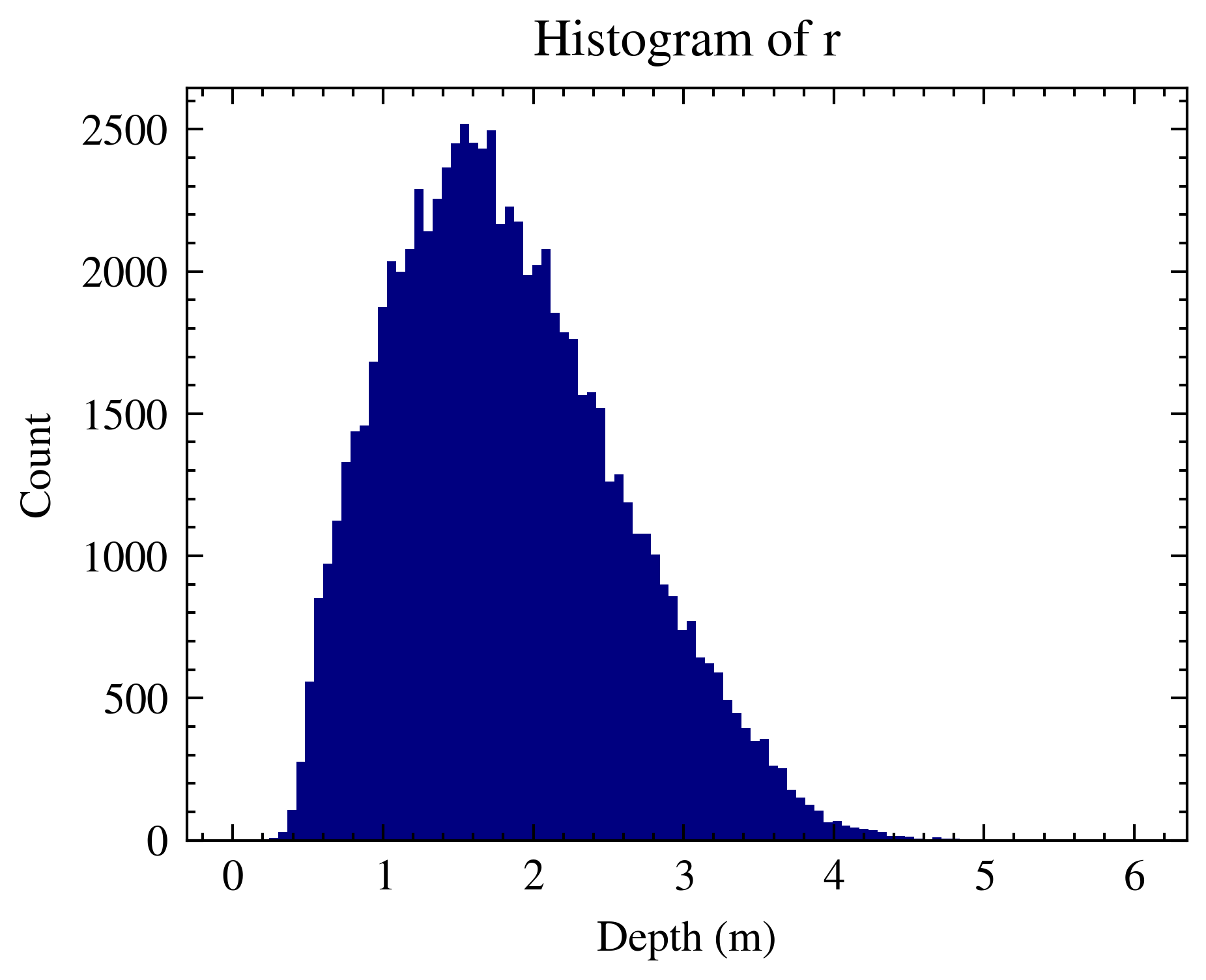}
        \caption{}
        \label{fig:depth histogram}
    \end{subfigure}
    \caption{\textbf{(a)} Dataset capture setup. \textbf{(b)} Distribution of radial distances between the robot and person in the dataset. %\patrick{the plots look a little too matplotlib-y. I use the python SciencePlots package which has some presets to make your plots look more legit.} 
    }
    \label{fig: hardware setup}
\end{figure}

To label $r$, participants wore a hat with ArUco markers~\cite{garrido2014automatic}. Patches of the 360\degree{} RGB frame were re-projected using a pinhole camera model and an ArUco pose estimator was run on these frames. Due to factors such as motion blur, lighting, and low resolution at further distances, ArUco markers were detected in only 74\% of all the frames in the dataset. The distribution of $r$ is shown in Fig.~\ref{fig:depth histogram}. Each dataset sample consists of a 1s clip of audio and corresponding video. We use the first frame of the video clip to extract location labels. We sample overlapping clips at 4Hz. We note that while robust RGB-D cameras are readily available, they are limited to a narrow field of view and not suitable for our 360\degree{} data.

 \subsection{Hardware}
 
 All audio recordings were done at 44.1 kHz using 4 RØDE NT2-A microphones connected to a MOTU M6 audio interface. The polar pattern for all 4 microphones was set to cardioid with the front side facing outwards. To ensure that we capture the maximum amount of information from the audio, the microphone gain was adjusted to the highest setting, the PAD was set to 0 dB and the high-pass filter was set to flat. An Insta360 ONE X2 was used for the 360\degree{} video recordings, with the video frames processed into 1440x720 resolution using equirectangular projection. The hardware was mounted on the Stretch RE-1 robot (Fig.~\ref{fig:dataset third person}). For the Empty Augmentation dataset, a standard tripod was used.

\begin{figure*}[t!]
    \centering
    \includegraphics[width=0.98\textwidth]{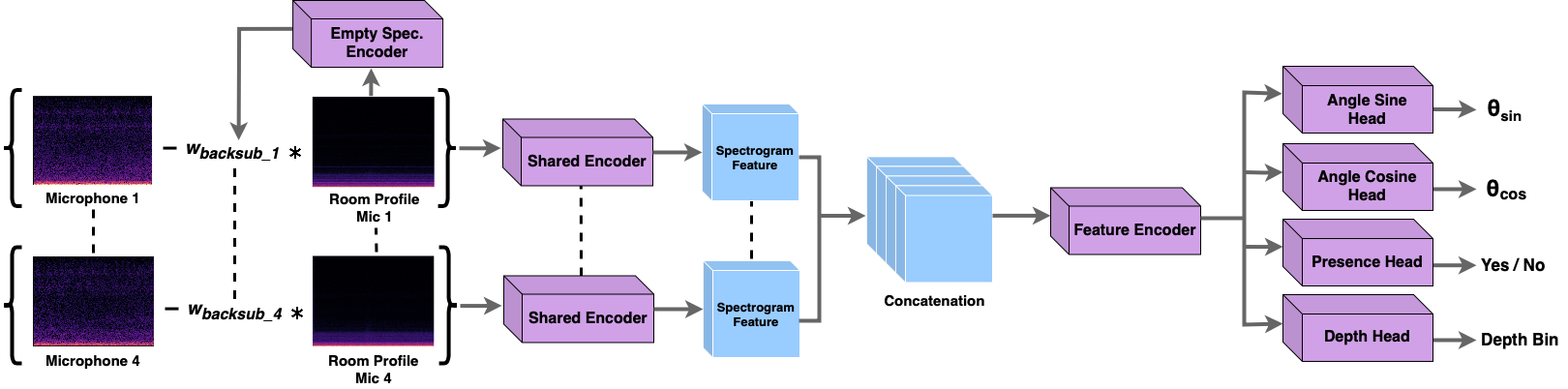}
    \caption{Diagram of our model architecture. We perform background subtraction (Sec.~\ref{sec:back sub}) on input spectrograms before passing them through a spectrogram encoder with shared weights. The resulting features are concatenated and passed through the feature encoder based on the ASPP module~\cite{chen2018encoder}. The output is fed to 4 linear layer heads for the prediction tasks.}
    \label{fig:architecture}
\end{figure*}

\section{Methodology}

We examine acoustic localization of people using only the incidental sounds produced by their moving presence. We design and train models on our dataset of high-quality multi-channel audio paired with 360\degree{} RGB data from which we extract location labels. While we collected data of people standing still as well, our paper only analyzes the moving actions (quiet, normal, loud). We train our models using leave-one-out cross validation across all 8 rooms in the dataset. All results are from averaging test performance across the 8 unseen rooms.

\subsection{Background Subtraction} \label{sec:back sub}

Given the weak-signal nature of the data, our models must be robust to the ambient noises in different rooms. As seen in Fig.~\ref{fig:spectrograms}, aside from loud walking, the other actions are difficult to distinguish from that of an empty room. This shows that most of the audio signals consist of background noise that our model must learn to ignore. To do so, we use a simple form of spectral subtraction. Right before recording in each room, we first collected 20s of empty room audio with either a static or dynamic robot. We split that audio into non-overlapping 1s clips, compute their spectrograms, and then take the average spectrogram. This average spectrogram, $S_{empty}$, is the empty room profile. Now given an input spectrogram of audio from the same room, $S_{in}$, we perform \textit{background subtraction} by computing $S_{final} = S_{in} - w_{backsub} * S_{empty}$. $w_{backsub}$ is a scalar weight that the empty spectrogram encoder learns (Fig.~\ref{fig:architecture}). We clamp the values to [0, 1].

\subsection{Empty Room Augmentation} \label{sec:empty aug}

Training our models to adapt to different background noises requires a wide variety of rooms to be seen during training. To supplement the 8 rooms we collected data in, we synthetically create additional rooms during training. Since audio is additive, we can inject additional noise into an audio recording and place the sounds of a moving person in a new room, one which contains a combination of ambient noises from two different rooms. This is the purpose of the Empty Augmentation dataset from Section~\ref{sec: empty room}. Given an audio clip from the \datasetName dataset, $x_r(t)$, and an audio clip of equal length of an empty room from the Empty Augmentation dataset, $x_{aug}(t)$, we first normalize both clips to an RMS of $0.02$. Then, we calculate a linear combination of the two waveforms: $x_{syn}(t) = (1 - w_{aug}) * x_r(t) + w_{aug} * x_{aug}(t)$. $w_{aug}$ is a scalar which we tune as a hyperparameter. We normalize $x_{syn}(t)$ again before processing its spectrogram.

For background subtraction, the same procedure described in Section~\ref{sec:back sub} is used to calculate empty room profiles for both the natural and synthetic empty room, resulting in $S_{empty}^{nat}$  and $S_{empty}^{syn}$. To compute the final empty room profile, the same weighting factor $w_{aug}$ is used: $S_{empty} = (1 - w_{aug}) * S_{empty}^{nat} + w_{aug} * S_{empty}^{syn}$. $S_{empty}$ is then used for background subtraction.

\subsection{Models}

\textbf{Architecture:} We adapt the audio encoder architecture from Vasudevan et al.~\cite{vasudevan2020semantic} for our person detection task. The network takes a 1s clip of audio in the form of a spectrogram. To generate the spectrogram, we first normalize the raw waveform to a constant RMS value of $0.02$. The waveform is then fed through a Short-Time Fourier Transform (STFT) with a window size of 512 and a hop length of 128 and then converted to the log scale. This results in a $[2, 257, 345]$ spectrogram for each microphone, where the 2 channels correspond to the real and complex components. 

As seen in the architecture diagram in Fig.~\ref{fig:architecture}, we first subtract the empty room profile spectrogram from the input spectrogram (Section~\ref{sec:back sub}) for each microphone before passing them individually through a spectrogram encoder with shared weights, consisting of 4 strided convolutional layers. The output is a $[256, 60, 120]$ feature map for each spectrogram. These features are then concatenated in the channel dimension to form a $[256n, 60, 120]$ feature map, where $n$ is the number of microphones being used, before being passed through the feature encoder. The feature encoder is an Atrous Spatial Pyramid Pooling (ASPP) module~\cite{chen2018encoder}, which~\cite{vasudevan2020semantic} found to be a powerful audio encoder for spatial audio tasks. We refer readers to~\cite{chen2018encoder, vasudevan2020semantic} for a more detailed description of the architecture. The output of the feature encoder is a $[1, 240, 480]$ feature map. The flattened feature map is passed through 4 task-specific decoders, each being a linear layer. Each decoder predicts one of the following: 

\textbf{Azimuthal angle prediction}: $\hat{\theta}_{sin}$ and $\hat{\theta}_{cos}$ are each predicted by a decoder. We clamp the predictions to $[-1, 1]$ and then decode back to the pixel coordinate $\hat{x} = tan^{-1}(\hat{\theta}_{sin} / \hat{\theta}_{cos})$. We then apply a L1 loss between $\hat{x}$ and $x$. The loss values for empty room training samples are ignored so the model does not try to learn to predict the location of a non-existent person.

\textbf{Radial distance prediction}: We frame radial distance $r$ estimation as a binary classification task by predicting if a person is within 1.7m of the robot, which is the median of the distribution in Fig.~\ref{fig:depth histogram}. We train using a binary cross-entropy loss and losses for empty room samples are ignored.

\textbf{Motion presence prediction}: The model learns a binary classification task of whether or not a person is moving in the room. A binary-cross entropy loss is used.

\textbf{Training:} We use the Adam~\cite{kingma2017adam} optimizer with a learning rate of $10^{-4}$, momentum of $0.9$, and weight decay of $10^{-3}$. We train the entire model using a multi-task framework on a single NVIDIA A40 GPU. The model has 8.37M parameters.

\begin{figure}[t!]
    \centering
    \includegraphics[width=0.98\columnwidth]{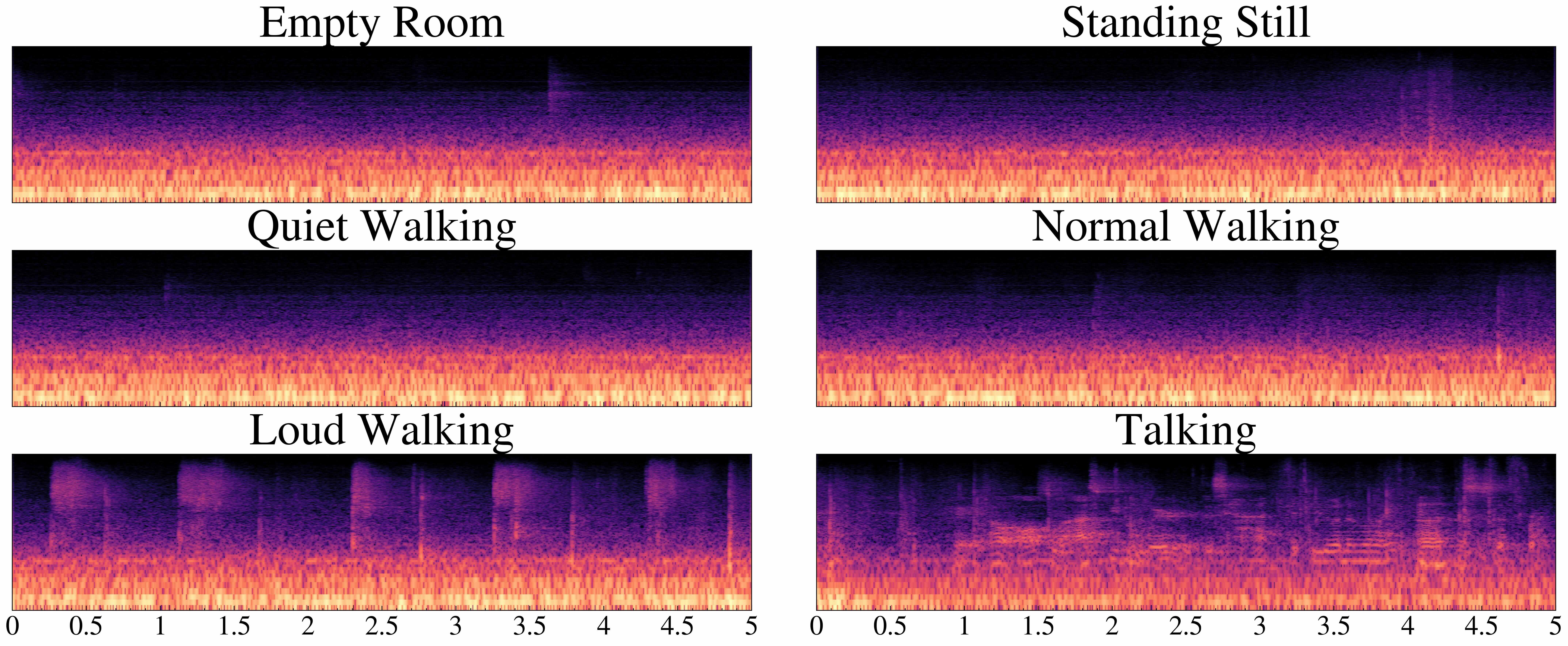}
    \caption{Log spectrograms for all categories along with regular talking. No talking is used in our work, but we show the spectrogram as reference for a common sound source used in localization. All recordings were taken in the same room during the same recording session.}
    \label{fig:spectrograms}
\end{figure}

\section{Experiments}

We present the person detection performance of our model trained on the \datasetName dataset. Performance is broken down into 3 tasks: $\theta$, $r$, and moving presence prediction. All results are the average performance across the 8 test room folds. We then compare against other methods and perform ablation studies to validate our model design. Finally, we demonstrate our model in the real world by implementing it on the Stretch RE-1 robot.

\subsection{Model Comparisons}
For the angle prediction, we compare our model with GCC-PHAT~\cite{knapp1976generalized}, a commonly used handcrafted feature, and StereoCRW~\cite{chen2022sound}, an unsupervised method that learns spectrogram representations. Both methods estimate the time delay between two stereo channels from which the DOA can be calculated. For StereoCRW, we run inference with the provided pre-trained model weights which had been trained on significantly more data and demonstrates generalization capabilities. Both comparison models are designed for 2 microphones which can only predict the direction of sound within the range [-90\degree{}, 90\degree{}]. To compare with our 360\degree{} method, we use an oracle which always selects the pair of microphones (front 2 or back 2 microphones) facing the person. We also compare against a naive oracle method, constant front, which always predicts 0\degree{} (straight ahead) relative to the microphone pair selected by the oracle.

Both radial distance and moving presence prediction are binary classification tasks, which we compare against chance. While Chen et al.~\cite{chen2021structure} examines a related task of estimating distance to nearby walls based on ambient sounds of the room, they focus on non-sound producing objects. Meanwhile, we treat ambient sounds as noise and instead focus on the subtle sounds that are present within.

\subsection{Azimuthal Angle Prediction} \label{sec:anglepred}

\begin{table}[t]
\caption{Mean absolute error (MAE) in degrees for azimuthal angle prediction of our model and comparison methods across the 3 actions divided by static (Sta.) and dynamic (Dyn.) robot. Our Base 4 Mics model is trained without background subtraction (Sec.~\ref{sec:back sub}) and empty room augmentation (Sec.~\ref{sec:empty aug}).}
\label{table:mae}
\begin{center}
\renewcommand{\arraystretch}{1.3}
\resizebox{\columnwidth}{!}{%
\begin{tabular}{cccccccc}
 % &&\multicolumn{3}{c}{\textbf{\textsc{Movement Categories}}} \\
\hline
& & \multicolumn{2}{c}{\textit{Quiet}} & \multicolumn{2}{c}{\textit{Normal}} & \multicolumn{2}{c}{\textit{Loud}} \\
\textsc{Category} & \textsc{Model} & Sta. & Dyn. & Sta. & Dyn. & Sta. & Dyn. \\
\hline
\hline
Random & Uniform 360\degree{} & 90 & 90 & 90 & 90 & 90 & 90 \\
\hdashline
\multirow{3}{*}{Oracle Mic Pair} & Constant Front & 50 & 43 & 50 & 46 & 50 & 43 \\
& GCC-PHAT~\cite{knapp1976generalized} & 44 & 47 & 45 & 43 & 46 & 47 \\
& StereoCRW~\cite{chen2022sound} & 52 & 46 & 51 & 48 & 37 & 34 \\
\hdashline
\multirow{4}{*}{Ours} & 1 Mic & 67 & 75 & 64 & 71 & 64 & 74 \\
& 2 Mics & 37 & 54 & 37 & 48 & 36 & 47 \\
& Base 4 Mics & 47 & 55 & 50 & 48 & 49 & 47 \\
% & 4 Mics & \textbf{23} & \textbf{27} & \textbf{24} & \textbf{23} & \textbf{21} & \textbf{23} \\
& 4 Mics & \textbf{21} & \textbf{26} & \textbf{22} & \textbf{24} & \textbf{19} & \textbf{22}\\
\hline
\end{tabular}}
\end{center}
\end{table}

We evaluate 3 variations of our model, each trained on a different number of microphones: all 4 microphones, the front 2 microphones, and a single front microphone. We calculate the mean absolute error (MAE) in degrees to measure angle prediction performance. We show separate metrics for inference on static and dynamic robot recordings.

Looking at Table~\ref{table:mae}, we notice that both GCC-PHAT and StereoCRW have similar performance as the naive constant front method. This supports our claim that the subtle sounds we focus on are difficult for previous sound localization methods. An exception is StereoCRW on the loud category, which performs noticeably better than the other methods being compared. Fig.~\ref{fig:spectrograms} qualitatively shows how the acoustic characteristics of the loud category is similar to that of talking. Since other sound localization methods tend to focus on relatively prominent sources of sounds, it makes sense that a similar category like loud walking is detectable as well. But for the more subtle actions, other methods are unable to pick out the useful sounds from the background noise.

Moving on to our models' performance, our 4-microphone model significantly outperforms all other methods, with both GCC-PHAT and StereoCRW having approximately twice the MAE across all categories. This shows that the incidental sounds created by a moving person provides a rich source of cues for person detection. Performance on static robot recordings are generally better than dynamic robot, suggesting that the added self-noise of a moving robot complicates the already difficult task of detecting these subtle sounds.

Looking next at our 2-microphone model, it outperforms the other methods for all static recordings. Recall that our 2-microphone model is at a disadvantage compared to the non-random methods we compare to, since those models have access to all 4 microphones and always picks the ideal pair facing the person. They only need to predict angles within the range [-90\degree{}, 90\degree{}] while our model, with access to only a fixed pair of microphones, has to predict the entire 360\degree{} range. This again highlights our model's ability to detect and localize the subtle, incidental sounds produced by people when they move, even under constrained input settings. 

While methods using time difference estimation can only unambiguously predict the direction of sound within 180\degree{} due to symmetry, the configuration of our microphones allows us to predict a wider angle range. We postulate that the cardioid polar pattern on our microphones breaks this symmetry by being more sensitive to the sounds coming from the front of the microphone versus the back. The model is able to learn from this subtle difference and determine which side the sound is coming from, even if the time difference between the two microphones is the same.
% It is reasonable that the angle prediction performance degrades when our model is trained on fewer microphones. But an interesting observation is that the degradation in performance is worse for the dynamic robot recordings than the static robot. This supports our hypothesis that the 2-microphone model relies on subtle differences between front versus back. The added self-noise of the robot, which is often louder than the incidental sounds the participants produced when they move, makes it harder for the model to detect these useful signals. 
Finally, we also train our model on audio from only 1 fixed microphone. Understandably, this variation performs worse than the non-random methods. But the 1-microphone model still performs better than chance, suggesting that it is able to pick up some cues from the mono audio.

\begin{table}[t]
    \begin{center}
        \setlength{\tabcolsep}{8pt}
        \begin{subtable}[h]{\columnwidth}
        \centering
        \begin{tabular}{cccccc}
        % &\multicolumn{3}{c}{\textbf{\textsc{Movement Categories}}} \\
        \multicolumn{6}{c}{\textbf{Overall Accuracy: 67\%}}\\
        \hline
        \multicolumn{2}{c}{\textit{Quiet}} & \multicolumn{2}{c}{\textit{Normal}} & \multicolumn{2}{c}{\textit{Loud}}\\
        Sta. & Dyn. & Sta. & Dyn. & Sta. & Dyn. \\
        \hline
        \hline
        % 66 & 62 & 70 & 67 & 70 & 66 \\
        67 & 61 & 71 & 66 & 71 & 66 \\
        \hline
        \end{tabular}%}
       \caption{}
       \label{table:depth}
    \end{subtable}
    \end{center}
    \hfill
    \begin{center}
        \setlength{\tabcolsep}{8pt}
        \begin{subtable}[h]{0.95\columnwidth}
        \centering
        \begin{tabular}{cc|cccccc}
        \multicolumn{8}{c}{\textbf{Overall Accuracy: 87\%}}\\
        \hline
        \multicolumn{2}{c|}{\textbf{\textsc{Negative}}} & \multicolumn{6}{c}{\textbf{\textsc{Positive}}} \\
        % & \multicolumn{3}{c}{\textit{Static Robot / Dynamic Robot}} \\
        % \hline
        \multicolumn{2}{c|}{\textit{Empty}} & \multicolumn{2}{c}{\textit{Quiet}} & \multicolumn{2}{c}{\textit{Normal}} & \multicolumn{2}{c}{\textit{Loud}} \\
        Sta. & Dyn. & Sta. & Dyn. & Sta. & Dyn. & Sta. & Dyn. \\
        \hline
        \hline
         81 & 85 & 89 & 80 & 95 & 92 & 96 & 94 \\
        \hline
        \end{tabular}
        \caption{}
        \label{table:presence}
     \end{subtable}
    \end{center} 
     \caption{\textbf{(a)} Binary classification accuracy (\%) of predicting if a moving person's radial distance is above or below 1.7m. Results are separated by action and static (Sta.) and dynamic (Dyn.) robot recordings. Chance is 50\%. \textbf{(b)} Binary (negative vs positive) classification accuracy (\%) of predicting if there is a moving person in the room. We also separate results by action, static (Sta.), and dynamic (Dyn.) robot recordings within each class. Chance is 50\%.}
\end{table}

\subsection{Radial Distance Prediction}

% \begin{table}[ht]
% \caption{Accuracy (\%) of our method on the depth binary classification task, separated by movement category and robot variation.}
% \label{table:depth}
% \begin{center}
% \setlength{\tabcolsep}{8pt}
% \renewcommand{\arraystretch}{1.35}
% \begin{tabular}{cccc}
% & \multicolumn{3}{c}{\textbf{\textsc{Movement Categories}}} \\
% \hline
% \textsc{Robot Variation} & \textit{Sneaky} & \textit{Regular} & \textit{Loud} \\
% \hline
% \hline
% Static & $89$ & $95$ & $96$ \\
% Dynamic & $80$ & $93$ & $94$ \\
% \end{tabular}
% \end{center}
% \end{table}

We also estimate the radial distance $r$ which, combined with $\theta$, gives us the location of the person. We frame this as a binary classification problem by setting the threshold to the median (1.7m) of the distribution (Fig.~\ref{fig:depth histogram}) and predicting if the person is above or below that threshold. The model is trained on 4 microphones and performs better than chance (50\%), shown in Table~\ref{table:depth}. Quiet walking is the most difficult action for this task since it has the least amount of signal. However, performance seems to saturate at normal and loud walking, with similar performance between both actions. 

\subsection{Moving Presence Prediction}

We also want to detect the presence of a moving person in addition to localizing them. We evaluate our models on the binary classification task of differentiating between the audio of a person present and moving in the room (positive) and that of an empty room (negative) in both static and dynamic robot recordings. The model is trained on 4 microphones and performs significantly better than chance (Table~\ref{table:presence}). As expected, the louder actions have better performance, since there are more obvious signals for the model to detect.

\subsection{Ablation Studies}

% \begin{table}[h]
% \caption{Angle prediction mean absolute error (MAE) in degrees for our a base 4-microphone model without background subtraction and empty room augmentation, along with models with \textbf{either} one.}
% \label{table:ablation}
% \begin{center}
% \setlength{\tabcolsep}{10pt}
% \renewcommand{\arraystretch}{1.3}
% \begin{tabular}{cccc}
%  % &\multicolumn{3}{c}{\textbf{\textsc{Movement Categories}}} \\
% \hline
% &\multicolumn{3}{c}{\textit{(Static Robot / Dynamic Robot)}} \\
% \textsc{Model} & \textit{Quiet} & \textit{Normal} & \textit{Loud} \\
% \hline
% \hline
% 4 Mics Base Model & 47 / 55 & 50 / 48 & 49 / 47 \\
% +Background Subtraction & 56 / 55 & 54 / 54 & 54 / 55 \\
% +Empty Room Aug. & 69 / 70 & 71 / 73 & 71 / 74 \\
% \hline
% \end{tabular}
% \end{center}
% \end{table}

We validate our model design by training a base, 4-microphone model with neither background subtraction nor empty room augmentation, shown in Table~\ref{table:mae}. The base model has twice the MAE, demonstrating the necessity of these two features and the difficulty of the sounds we are learning from. Without additional methods to remove background noise, deep learning models have trouble detecting these subtle sounds. We do not decouple background subtraction and empty room augmentation because they complement each other, with the former allowing the model to adapt to different types of background noises while the latter provides more diverse training data for the model to actually learn the conditioning. Also, the 2- and 1-microphone models discussed in Section~\ref{sec:anglepred} can be seen as an ablation on the number of microphones. 

\subsection{Robotic Human Awareness}

\begin{figure}[t]
    \centering
    \includegraphics[width=\columnwidth]{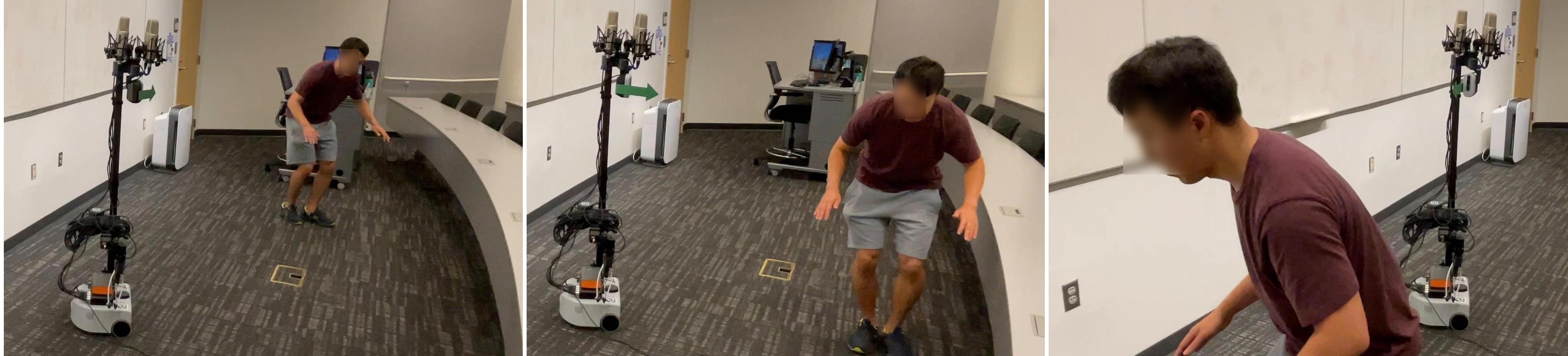}
    \caption{We implement our trained model on the Stretch RE-1 robot to track a person using only the incidental sounds created as they move quietly. The robot pans the RealSense camera, with green arrow attached, to face where the model estimates the person to be (zoom in for best results).}
    \label{fig:demo}
\end{figure}

We implement our trained model on the Stretch RE-1 to demonstrate robotic human awareness. Using the same hardware setup and in an unseen room, we input the most recent 1s clip of audio into the model and use the predicted angle to pan the RealSense camera on the Stretch to face the person. During the pan, the model does not perform inference to avoid the sound of the motor interfering with angle estimation. On a RTX 2080, the model runs at 142Hz. Fig.~\ref{fig:demo} provides a demonstration of the robot pointing at the person. We do not use the RealSense data in any way to estimate the person's direction. Given the narrow field of view of the RealSense (58\degree{}), we evaluate our algorithm's performance by determining the success rate of the person being present in the frame right after panning. We asked a participant to move at each of the 3 speeds for 4 minutes and obtained the following success rate: Quiet 80\%, Normal 79\%, Loud 82\%. 

\section{Conclusion}

We demonstrate the ability to localize people using only the subtle sounds they incidentally produce as they move. We present the Robot Kidnapper dataset and the resulting person detection models that can be implemented on robots to track a person as they move quietly. Our work opens up an avenue of exploration for how robots can learn human awareness with only passive audio sensing and without nearby humans needing to intentionally produce additional sounds.
 %without any additional involvement or participation from humans.

\textbf{Limitations:} 
Our dataset only contains instances of a single person in a room up to a maximum distance of 6m. We also have not tested our method on different types of microphones or robot setups. Additionally, we are unable to localize a person if they are standing completely still.

\bibliographystyle{unsrt}
\bibliography{references}

\begin{thebibliography}{10}

\bibitem{ren2015faster}
Shaoqing Ren, Kaiming He, Ross Girshick, and Jian Sun.
\newblock Faster r-cnn: Towards real-time object detection with region proposal networks.
\newblock In C.~Cortes, N.~Lawrence, D.~Lee, M.~Sugiyama, and R.~Garnett, editors, {\em Advances in Neural Information Processing Systems}, volume~28. Curran Associates, Inc., 2015.

\bibitem{he2017mask}
Kaiming He, Georgia Gkioxari, Piotr Doll{\'a}r, and Ross Girshick.
\newblock Mask r-cnn.
\newblock In {\em Proceedings of the IEEE international conference on computer vision}, pages 2961--2969, 2017.

\bibitem{carion2020end}
Nicolas Carion, Francisco Massa, Gabriel Synnaeve, Nicolas Usunier, Alexander Kirillov, and Sergey Zagoruyko.
\newblock End-to-end object detection with transformers.
\newblock In {\em European conference on computer vision}, pages 213--229. Springer, 2020.

\bibitem{Linder2021CrossModalAO}
Timm Linder, Narunas Vaskevicius, Robert Schirmer, and Kai~Oliver Arras.
\newblock Cross-modal analysis of human detection for robotics: An industrial case study.
\newblock {\em 2021 IEEE/RSJ International Conference on Intelligent Robots and Systems (IROS)}, pages 971--978, 2021.

\bibitem{jia2021domain}
Dan Jia, Alexander Hermans, and Bastian Leibe.
\newblock Domain and modality gaps for lidar-based person detection on mobile robots.
\newblock {\em arXiv e-prints}, pages arXiv--2106, 2021.

\bibitem{jia2020dr}
Dan Jia, Alexander Hermans, and Bastian Leibe.
\newblock Dr-spaam: A spatial-attention and auto-regressive model for person detection in 2d range data.
\newblock In {\em 2020 IEEE/RSJ International Conference on Intelligent Robots and Systems (IROS)}, pages 10270--10277. IEEE, 2020.

\bibitem{yan2018multisensor}
Zhi Yan, Li~Sun, Tom Duckctr, and Nicola Bellotto.
\newblock Multisensor online transfer learning for 3d lidar-based human detection with a mobile robot.
\newblock In {\em 2018 IEEE/RSJ International Conference on Intelligent Robots and Systems (IROS)}, pages 7635--7640, 2018.

\bibitem{fung2023robots}
Angus Fung, Beno Benhabib, and Goldie Nejat.
\newblock Robots autonomously detecting people: A multimodal deep contrastive learning method robust to intraclass variations.
\newblock {\em IEEE Robotics and Automation Letters}, 2023.

\bibitem{garrido2014automatic}
Sergio Garrido-Jurado, Rafael Mu{\~n}oz-Salinas, Francisco~Jos{\'e} Madrid-Cuevas, and Manuel~Jes{\'u}s Mar{\'\i}n-Jim{\'e}nez.
\newblock Automatic generation and detection of highly reliable fiducial markers under occlusion.
\newblock {\em Pattern Recognition}, 47(6):2280--2292, 2014.

\bibitem{chen2018encoder}
Liang-Chieh Chen, Yukun Zhu, George Papandreou, Florian Schroff, and Hartwig Adam.
\newblock Encoder-decoder with atrous separable convolution for semantic image segmentation.
\newblock {\em arXiv:1802.02611}, 2018.

\bibitem{Linder2018TowardsA3}
Timm Linder, Dennis Grie{\ss}er, Narunas Vaskevicius, and Kai~Oliver Arras.
\newblock Towards accurate 3d person detection and localization from rgb-d in cluttered environments.
\newblock 2018.

\bibitem{Linder2020accurate}
T.~{Linder}, K.~Y. {Pfeiffer}, N.~{Vaskevicius}, R.~{Schirmer}, and K.~O. {Arras}.
\newblock Accurate detection and {3D} localization of humans using a novel {YOLO}-based {RGB-D} fusion approach and synthetic training data.
\newblock In {\em 2020 IEEE International Conference on Robotics and Automation (ICRA)}, 2020.

\bibitem{beyer2018deep}
Lucas Beyer, Alexander Hermans, Timm Linder, Kai Arras, and Bastian Leibe.
\newblock Deep person detection in 2d range data.
\newblock {\em IEEE Robotics and Automation Letters}, PP, 04 2018.

\bibitem{2021arXiv210611239J}
Dan {Jia}, Alexander {Hermans}, and Bastian {Leibe}.
\newblock {2D vs. 3D LiDAR-based Person Detection on Mobile Robots}.
\newblock {\em arXiv e-prints}, page arXiv:2106.11239, June 2021.

\bibitem{ku2018joint}
Jason Ku, Melissa Mozifian, Jungwook Lee, Ali Harakeh, and Steven~L. Waslander.
\newblock Joint 3d proposal generation and object detection from view aggregation.
\newblock page 1–8. IEEE Press, 2018.

\bibitem{linder2016on}
Timm Linder, Stefan Breuers, Bastian Leibe, and Kai~O. Arras.
\newblock On multi-modal people tracking from mobile platforms in very crowded and dynamic environments.
\newblock In {\em 2016 IEEE International Conference on Robotics and Automation (ICRA)}, pages 5512--5519, 2016.

\bibitem{singh2023sporadic}
Gaurav Singh, Paul Ghanem, and Taskin Padir.
\newblock Sporadic audio-visual embodied assistive robot navigation for human tracking.
\newblock In {\em Proceedings of the 16th International Conference on PErvasive Technologies Related to Assistive Environments}, PETRA '23, page 99–105, New York, NY, USA, 2023. Association for Computing Machinery.

\bibitem{yamazaki2020audio}
Yuki Yamazaki, Chinthaka Premachandra, and Chamika~Janith Perea.
\newblock Audio-processing-based human detection at disaster sites with unmanned aerial vehicle.
\newblock {\em IEEE Access}, 8:101398--101405, 2020.

\bibitem{Ciuffreda2023people}
Ilaria Ciuffreda, Gianmarco Battista, Sara Casaccia, and Gian~Marco Revel.
\newblock People detection measurement setup based on a doa approach implemented on a sensorised social robot.
\newblock {\em Measurement: Sensors}, 25:100649, 2023.

\bibitem{sewtz2020robust}
Marco Sewtz, Tim Bodenmüller, and Rudolph Triebel.
\newblock Robust music-based sound source localization in reverberant and echoic environments.
\newblock In {\em 2020 IEEE/RSJ International Conference on Intelligent Robots and Systems (IROS)}, pages 2474--2480, 2020.

\bibitem{portello2011acoustic}
Alban Portello, Patrick Danès, and Sylvain Argentieri.
\newblock Acoustic models and kalman filtering strategies for active binaural sound localization.
\newblock In {\em 2011 IEEE/RSJ International Conference on Intelligent Robots and Systems}, pages 137--142, 2011.

\bibitem{bechler2004system}
D.~Bechler, M.S. Schlosser, and K.~Kroschel.
\newblock System for robust 3d speaker tracking using microphone array measurements.
\newblock In {\em 2004 IEEE/RSJ International Conference on Intelligent Robots and Systems (IROS) (IEEE Cat. No.04CH37566)}, volume~3, pages 2117--2122 vol.3, 2004.

\bibitem{shi2023audio}
Zhanbo Shi, Lin Zhang, and Dongqing Wang.
\newblock Audio-visual sound source localization and tracking based on mobile robot for the cocktail party problem.
\newblock {\em Applied Sciences}, 13(10), 2023.

\bibitem{Gala2018RealtimeAS}
Deepak Gala, Nathan Lindsay, and Liang Sun.
\newblock Realtime active sound source localization for unmanned ground robots using a self-rotational bi-microphone array.
\newblock {\em Journal of Intelligent \& Robotic Systems}, 95:935 -- 954, 2018.

\bibitem{knapp1976generalized}
C.~Knapp and G.~Carter.
\newblock The generalized correlation method for estimation of time delay.
\newblock {\em IEEE Transactions on Acoustics, Speech, and Signal Processing}, 24(4):320--327, 1976.

\bibitem{chen2022sound}
Ziyang Chen, David~F Fouhey, and Andrew Owens.
\newblock Sound localization by self-supervised time delay estimation.
\newblock {\em European Conference on Computer Vision (ECCV)}, 2022.

\bibitem{Tripathi2021human}
Abhinav Tripathi, Mohd.~Ammar Khan, Akash Pandey, Pankaj Yadav, and Amit~Kumar Sharma.
\newblock Human following robot using ultrasonic sensor.
\newblock In {\em 2021 3rd International Conference on Advances in Computing, Communication Control and Networking (ICAC3N)}, pages 764--770, 2021.

\bibitem{Christensen2019BatVisionLT}
Jesper~Haahr Christensen, Sascha Hornauer, and Stella~X. Yu.
\newblock Batvision: Learning to see 3d spatial layout with two ears.
\newblock {\em 2020 IEEE International Conference on Robotics and Automation (ICRA)}, pages 1581--1587, 2019.

\bibitem{wang2023soundcam}
Mason Wang, Samuel Clarke, Jui-Hsien Wang, Ruohan Gao, and Jiajun Wu.
\newblock Soundcam: A dataset for finding humans using room acoustics.
\newblock In {\em Advances in Neural Informaion Processing Systems}, 2023.

\bibitem{tan2020efficientdet}
Mingxing Tan, Ruoming Pang, and Quoc~V Le.
\newblock Efficientdet: Scalable and efficient object detection.
\newblock In {\em Proceedings of the IEEE/CVF conference on computer vision and pattern recognition}, pages 10781--10790, 2020.

\bibitem{zhou2022detecting}
Xingyi Zhou, Rohit Girdhar, Armand Joulin, Philipp Kr{\"a}henb{\"u}hl, and Ishan Misra.
\newblock Detecting twenty-thousand classes using image-level supervision.
\newblock In {\em European Conference on Computer Vision}, pages 350--368. Springer, 2022.

\bibitem{kirillov2023segment}
Alexander Kirillov, Eric Mintun, Nikhila Ravi, Hanzi Mao, Chloe Rolland, Laura Gustafson, Tete Xiao, Spencer Whitehead, Alexander~C Berg, Wan-Yen Lo, et~al.
\newblock Segment anything.
\newblock {\em arXiv preprint arXiv:2304.02643}, 2023.

\bibitem{premebida2014pedestrian}
Cristiano Premebida, Joao Carreira, Jorge Batista, and Urbano Nunes.
\newblock Pedestrian detection combining rgb and dense lidar data.
\newblock In {\em 2014 IEEE/RSJ International Conference on Intelligent Robots and Systems}, pages 4112--4117. IEEE, 2014.

\bibitem{navarro2016machine}
Pedro~J Navarro, Carlos Fernandez, Raul Borraz, and Diego Alonso.
\newblock A machine learning approach to pedestrian detection for autonomous vehicles using high-definition 3d range data.
\newblock {\em Sensors}, 17(1):18, 2016.

\bibitem{wilson2021argoverse}
Benjamin Wilson, William Qi, Tanmay Agarwal, John Lambert, Jagjeet Singh, Siddhesh Khandelwal, Bowen Pan, Ratnesh Kumar, Andrew Hartnett, Jhony~Kaesemodel Pontes, et~al.
\newblock Argoverse 2: Next generation datasets for self-driving perception and forecasting.
\newblock In {\em Thirty-fifth Conference on Neural Information Processing Systems Datasets and Benchmarks Track (Round 2)}, 2021.

\bibitem{yu2020scale}
Xuehui Yu, Yuqi Gong, Nan Jiang, Qixiang Ye, and Zhenjun Han.
\newblock Scale match for tiny person detection.
\newblock In {\em Proceedings of the IEEE/CVF winter conference on applications of computer vision}, pages 1257--1265, 2020.

\bibitem{park2019cnn}
Jisoo Park, Jingdao Chen, Yong~K Cho, Dae~Y Kang, and Byung~J Son.
\newblock Cnn-based person detection using infrared images for night-time intrusion warning systems.
\newblock {\em Sensors}, 20(1):34, 2019.

\bibitem{zin2011fusion}
Thi~Thi Zin, Hideya Takahashi, Takashi Toriu, and Hiromitsu Hama.
\newblock Fusion of infrared and visible images for robust person detection.
\newblock {\em Image fusion}, pages 239--264, 2011.

\bibitem{ye2021deep}
Mang Ye, Jianbing Shen, Gaojie Lin, Tao Xiang, Ling Shao, and Steven~CH Hoi.
\newblock Deep learning for person re-identification: A survey and outlook.
\newblock {\em IEEE transactions on pattern analysis and machine intelligence}, 44(6):2872--2893, 2021.

\bibitem{ristani2018features}
Ergys Ristani and Carlo Tomasi.
\newblock Features for multi-target multi-camera tracking and re-identification.
\newblock In {\em Proceedings of the IEEE conference on computer vision and pattern recognition}, pages 6036--6046, 2018.

\bibitem{object2015occlusion}
Samarth Brahmbhatt, Heni~Ben Amor, and Henrik Christensen.
\newblock Occlusion-aware object localization, segmentation and pose estimation.
\newblock In {\em Proceedings of the British Machine Vision Conference (BMVC)}, pages 80.1--80.13. BMVA Press, September 2015.

\bibitem{ceriani2013optimal}
Nicola~Maria Ceriani, Giovanni~Buizza Avanzini, Andrea~Maria Zanchettin, Luca Bascetta, and Paolo Rocco.
\newblock Optimal placement of spots in distributed proximity sensors for safe human-robot interaction.
\newblock In {\em 2013 IEEE International Conference on Robotics and Automation}, pages 5858--5863. IEEE, 2013.

\bibitem{zieger2009acoustic}
Christian Zieger, Alessio Brutti, and Piergiorgio Svaizer.
\newblock Acoustic based surveillance system for intrusion detection.
\newblock In {\em 2009 Sixth IEEE International Conference on Advanced Video and Signal Based Surveillance}, pages 314--319. IEEE, 2009.

\bibitem{chu2006scene}
Selina Chu, Shrikanth Narayanan, C-C~Jay Kuo, and Maja~J Mataric.
\newblock Where am i? scene recognition for mobile robots using audio features.
\newblock In {\em 2006 IEEE International conference on multimedia and expo}, pages 885--888. IEEE, 2006.

\bibitem{hornstein2006sound}
Jonas Hornstein, Manuel Lopes, Jos{\'e} Santos-Victor, and Francisco Lacerda.
\newblock Sound localization for humanoid robots-building audio-motor maps based on the hrtf.
\newblock In {\em 2006 IEEE/RSJ International Conference on Intelligent Robots and Systems}, pages 1170--1176. IEEE, 2006.

\bibitem{liang2019making}
Hongzhuo Liang, Shuang Li, Xiaojian Ma, Norman Hendrich, Timo Gerkmann, Fuchun Sun, and Jianwei Zhang.
\newblock Making sense of audio vibration for liquid height estimation in robotic pouring.
\newblock In {\em 2019 IEEE/RSJ International Conference on Intelligent Robots and Systems (IROS)}, pages 5333--5339. IEEE, 2019.

\bibitem{chen2021structure}
Ziyang Chen, Xixi Hu, and Andrew Owens.
\newblock Structure from silence: Learning scene structure from ambient sound.
\newblock In {\em 5th Annual Conference on Robot Learning}, 2021.

\bibitem{valin2003robust}
J-M Valin, Fran{\c{c}}ois Michaud, Jean Rouat, and Dominic L{\'e}tourneau.
\newblock Robust sound source localization using a microphone array on a mobile robot.
\newblock In {\em Proceedings 2003 IEEE/RSJ International Conference on Intelligent Robots and Systems (IROS 2003)(Cat. No. 03CH37453)}, volume~2, pages 1228--1233. IEEE, 2003.

\bibitem{wilson2020avot}
Justin Wilson and Ming~C Lin.
\newblock Avot: Audio-visual object tracking of multiple objects for robotics.
\newblock In {\em 2020 IEEE International Conference on Robotics and Automation (ICRA)}, pages 10045--10051. IEEE, 2020.

\bibitem{sasaki2018online}
Yoko Sasaki, Ryo Tanabe, and Hiroshi Takernura.
\newblock Online spatial sound perception using microphone array on mobile robot.
\newblock In {\em 2018 IEEE/RSJ International Conference on Intelligent Robots and Systems (IROS)}, pages 2478--2484. IEEE, 2018.

\bibitem{vasudevan2020semantic}
Arun~Balajee Vasudevan, Dengxin Dai, , and Luc {Van Gool}.
\newblock Semantic object prediction and spatial sound super-resolution with binaural sounds.
\newblock {\em ECCV}, 2020.

\bibitem{kingma2017adam}
Diederik~P. Kingma and Jimmy Ba.
\newblock Adam: A method for stochastic optimization, 2017.

\end{thebibliography}

\end{document}